\pdfoutput=1
\documentclass[11pt]{article}

\usepackage{EMNLP2023}

\usepackage{times}
\usepackage{latexsym}
\usepackage{placeins}

\usepackage[T1]{fontenc}
\usepackage[utf8]{inputenc}

\usepackage{microtype}

\usepackage{inconsolata}

\usepackage{amsmath}
\usepackage{amsfonts}
\usepackage{hhline}
\usepackage{multirow}
\usepackage[T1]{fontenc}    % use 8-bit T1 fonts
\usepackage{url}            % simple URL typesetting
\usepackage{booktabs, multirow, tabularx}       % professional-quality tables
\usepackage{amsfonts}       % blackboard math symbols
\usepackage{nicefrac}       % compact symbols for 1/2, etc.
\usepackage{microtype}      % microtypography
\usepackage{graphicx}
\usepackage{natbib}
\usepackage{doi}
\usepackage{color}
\usepackage{amsmath}
\usepackage{mathtools, cuted}
\usepackage{etoolbox}
\usepackage{changepage}
\usepackage{pdflscape}
\usepackage{multicol}
\usepackage{balance}
\usepackage{wrapfig}
\usepackage{array}
\usepackage{soul}
\usepackage{bm}
\usepackage{enumitem}

\AfterEndEnvironment{strip}{\leavevmode}

\newcommand{\nce}{\mathrm{NCE}}

\newcommand{\JBert}{\texttt{Jina\,BERT}}

\newcommand{\rom}[1]{\uppercase\expandafter{\romannumeral #1\relax}}
\newcommand{\MultilingualEFiveBase}{\texttt{multilingual-e5-base}}
\newcommand{\JinaDe}{\texttt{jina-de-base}}

\newcommand{\JinaEs}{\texttt{jina-es-base}}

\definecolor{darkgreen}{rgb}{0.0, 0.5, 0.1}

\title{Multi-Task Contrastive Learning for 8192-Token Bilingual Text Embeddings}
\author{Isabelle Mohr, Markus Krimmel, Saba Sturua, Mohammad Kalim Akram, \\\textbf{Andreas Koukounas}, \textbf{Michael G\"unther}, \textbf{Georgios Mastrapas}, \textbf{Vinit Ravishankar},\\ \textbf{Joan Fontanals Martínez}, \textbf{Feng Wang}, \textbf{Qi Liu},  \textbf{Ziniu Yu}, \textbf{Jie Fu}, \textbf{Saahil Ognawala}, \\\textbf{Susana Guzman}, \textbf{Bo Wang}, \textbf{Maximilian Werk}, \textbf{Nan Wang} \and \textbf{Han Xiao}\\
	Jina AI GmbH, Ohlauer Str. 43, 10999 Berlin, Germany \\
	\texttt{research@jina.ai}}
\date{2024/02/26}

\hypersetup{
pdftitle={Multi-Task Contrastive Learning for 8192-Token Bilingual Text Embeddings},
pdfauthor={Isabelle Mohr, Markus Krimmel, Saba Sturua, Mohammad Kalim Akram, Andreas Koukounas, Michael G\"unther, Georgios Mastrapas, Vinit Ravishankar, Joan Fontanals Martínez, Feng Wang, Qi Liu, Ziniu Yu, Jie Fu, Saahil Ognawala, Susana Guzman, Bo Wang, Maximilian Werk, Nan Wang, Han Xiao},
pdfkeywords={Embeddings, Bilingual Models, Multilingual Models, Multi-Task Learning, Large Language Models, LLM, 8K, OpenAI, BERT, Token length, Semantic Textual Similarity},
}

\begin{document}
\maketitle
\begin{abstract}

We introduce a novel suite of state-of-the-art bilingual text embedding models that are  designed to support English and another target language. These models are capable of processing lengthy text inputs with up to 8192 tokens, making them highly versatile for a range of natural language processing tasks such as text retrieval, clustering, and semantic textual similarity (STS) calculations.

By focusing on bilingual models and introducing a unique multi-task learning objective, we have significantly improved the model performance on STS tasks, which outperforms the capabilities of existing multilingual models in both target language understanding and cross-lingual evaluation tasks. Moreover, our bilingual models are more efficient, requiring fewer parameters and less memory due to their smaller vocabulary needs. Furthermore, we have expanded the Massive Text Embedding Benchmark (MTEB) to include benchmarks for German and Spanish embedding models. This integration aims to stimulate further research and advancement in text embedding technologies for these languages.

\end{abstract}

% Ideas for the paper:
% \begin{itemize}
%     \item Short description of the model training (referencing previous paper)
%     \item Multi-Task constrastive learning approach (STS-loss for STS datasets and infonce for triplets) -> also mention that we experimenting further with it for clustering
%     \item Why bilingual -> conduct an experiment (ablation study) of training different backbones on a small contrastive pair dataset and evaluated against MTEB -> proof that bilingual makes sense
%     \item Providing new evaluation benchmark for German and Spanish
%     \item (Ablation study for multi-task learning - compare using InfoNCE, PearsonLoss, and CosineLoss for STS training data -> Ask Vinit)
% \end{itemize}

\section{Introduction}
\label{sec:introduction}
Text embedding models allow NLP practitioners to encode a text snippet of arbitrary length into a vector in the semantic space.
Such embedding models have evolved into core components within the domains of neural information retrieval and found wide adoption and integration into various other facets of Natural Language Processing research and applications.
In particular, they became specifically popular for building Retrieval-Augmented Generation pipelines.
% The capabilities of embedding models to represent textual content in a structured numeric manner has led to their wide adoption and integration into various facets of natural language processing research and applications, underpinning the advancement of both information retrieval and text generation systems~\cite{gunther2023jina2}.

Currently, such dense text embedding models fall into two categories: monolingual and multilingual.
Monolingual models are usually limited to processing English text exclusively.
In contrast, most of the multilingual text embedding models support tens of languages.
This is achieved by fine-tuning pre-existing multilingual backbones like XLM-RoBERTa~\cite{conneau2020xlmr} on mainly English data.\footnote{e.g., multilingual-e5 is fine-tuned with > 99\% English data:  \url{https://huggingface.co/intfloat/multilingual-e5-large}}
% and M3~\cite{bge-m3} % m3 is also based on XLM roberta
Alternatively, multilingual knowledge distillation can be applied to align embedding models across various languages using parallel data~\cite{reimers2020making} to cope with the scarcity of high-quality semantic pairs or triplets in the target languages.
Despite these efforts, training data with such an extremely unbalanced distribution between different languages leads to inferior performance in the target language.
Moreover, compared to monolingual English models, the performance of multilingual models falls short in the English tasks, i.e., the multilingual version of the E5 model~\cite{wang2022text} achieves a significantly lower average score compared to its monolingual variant on the Massive Text Embedding Benchmark (MTEB)~\cite{muennighoff2023mteb}.
Furthermore, to support tens of languages, most of the multilingual models have a large token vocabulary. In practice, only two or three languages are used in the industry use cases, which means the parameters due to the large vocabulary are never used.

To address these shortcomings of multilingual models, we introduce a new family of embedding models namely~\textit{bilingual text embeddings}, that focus on handling two languages effectively.
Each of our bilingual language models consistently supports a target language (e.g., German, Chinese, or Spanish) alongside English.
Instead of relying on existing multilingual backbone models, we train backbone models that support these specific language pairs, thus reducing the model size compared to multilingual alternatives.
We show that these models can be fine-tuned more effectively on embedding tasks than multilingual models.
For fine-tuning the models on the text-embedding tasks, we introduce a novel multi-task learning strategy that increases the performance of the embedding models on STS and retrieval tasks.
Additionally, our bilingual models support large context lengths (up to 8192 tokens), as we are utilizing the same architecture used for our long context length model~\cite{gunther2023jina2}.
As there is only limited work done in evaluating embedding models in German and Spanish, we collect evaluation datasets for integrating them into the existing Massive Text Embedding Benchmark (MTEB) to build up a standard for benchmarking text embedding models in these languages.

Our efforts result in a suite of bilingual text embedding models\footnote{de-en:\url{https://huggingface.co/jinaai/jina-embeddings-v2-base-de}\\zh-en:\url{https://huggingface.co/jinaai/jina-embeddings-v2-base-zh}\\es-en:\url{https://huggingface.co/jinaai/jina-embeddings-v2-base-es}\\API: \url{https://jina.ai/embeddings}} that achieve superior or comparable performance on various embedding tasks compared to multilingual state-of-the-art models and specifically excel in cross-lingual retrieval settings while retaining a significantly smaller model size.

Section~\ref{sec:related_work} starts with an exploration of related research on training embedding models, multilingual models, and multi-task learning.
We continue with an overview of the training (Section~\ref{sec:training-overview}), a description of the pre-training phase (Section~\ref{sec:backbone_pretraining}), and the fine-tuning phase (Section~\ref{sec:embedding-training}).
In the evaluation in Section~\ref{sec:evaluation}, we compare the bilingual models to other state-of-the-art models and conduct ablation studies of the backbone training and the multi-task learning.
Finally, we draw conclusions in Section~\ref{sec:conclusion}.
\section{Related Work}
\label{sec:related_work}

While lots of research focuses on embedding training techniques, little research has been conducted on training bilingual embedding models.
Some research papers focus on multilingual models, for example, \citet{reimers2020making} propose a method to distill monolingual teacher embedding models to multilingual student language models and the embedding training technique proposed for the E5 model~\cite{wang2022text} has been successfully transferred to train a multilingual E5 model\footnote{\url{https://huggingface.co/intfloat/multilingual-e5-large}} that achieves better performance than many monolingual models of the same size on non-English tasks of the MTEB benchmark~\cite{muennighoff2023mteb}.
In a more recent work, \cite{bge-m3}, the authors propose M3, a single model for dense retrieval, multi-vector retrieval, and sparse retrieval for multiple languages.

\subsection{Multilingual Language Models}

For training a bilingual embedding model, we use a backbone language model that can handle multiple languages.
One of the first multilingual transformer models is the multilingual BERT (mBERT) model developed by \citet{devlin2019bert}, which is trained for 104 languages.
% There has been increasing interest in multilingual masked language modeling. \citet{devlin2019bert} released multilingual BERT (mBERT), trained on 104 languages. 
XLM~\citep{conneau2019xlm} and XLM-R~\citep{conneau2020xlmr} continue this line of work, leveraging parallel data during pre-training in their translation language modeling objective.
Various authors note difficulties in training massively multilingual models, owing to a phenomenon called \emph{the curse of multilinguality}~\citep{conneau2020xlmr}:
while some level of multilinguality leads to positive transfer between languages, model capacity becomes "diluted" across languages, eventually leading to performance degradations~\citep{conneau2020xlmr,arivazhagan2019capacitydilution,blevins2022pretraining_dynamics,wang2020interference}.
Some works address this issue via language-specific network components~\citep{pfeiffer2022xmod, pfeiffer2020madx}, optimized per-language vocabulary capacity~\citep{zheng2021vocap}, and a temporary capacity increase during pre-training~\citep{chung2021rembert}.
However, other authors argue in favor of tri- and bilingual models that can significantly outperform their multilingual counterparts~\citep{Xu2021bibert,chang2020finnish,ulcar2020moreisless}.

\subsection{Multi-Task Learning}

For training a general-purpose embedding model, we adopt the concept of multi-task learning~\citep{caruana1997multitask, crawshaw2020multitask-survey}.
Multi-task learners use shared representations to solve several related downstream tasks at once.
% Multi-task learners~\citep{caruana1997multitask, crawshaw2020multitask-survey} use shared representations to solve several related downstream tasks at once.
Since they share model parameters across tasks, a training signal from diverse task objectives may lead to inductive transfers~\citep{baxter2000model}. Early relevant works in this direction include~\citet{subramanian2018learning}, where sentence representations were learned in a multi-task fashion over a range of tasks, and~\citet{luong2015multi}, where multiple sequence-to-sequence tasks were learned simultaneously. 

Within this paradigm, therefore, parameters may be shared either in a \emph{hard} fashion through common network components or \emph{softly} via an objective that incentivizes the similarity of task-specific networks~\citep{duong2015soft}.
The overall loss of the multi-task network is usually formulated as a weighted sum of task losses, where the weights may change dynamically during training to emphasize underperforming tasks~\citep{guo2018task-performance}, balance learning speeds between tasks~\citep{chen2018gradnorm} or maximize the observed data likelihood when taking into account per-task uncertainty~\citep{kdendall2018uncertainty}. In our work, we adopt a hard parameter-sharing approach but deviate from the usual loss formulation. Instead of forming the overall loss as a sum of task losses, we sample a task in each forward pass and backpropagate the corresponding task loss.

\section{Training Paradigm Overview}
\label{sec:training-overview}

% \todo{adjust this section}

% The training paradigm for \JEmbeddingVTwo~is divided into three stages:
For training each of the bilingual embedding models, we follow a procedure which consists of three stages:

\begin{enumerate}[label=\Roman*]
\item \textbf{Pre-training a JinaBERT model:} The backbone model has the same implementation as the \JBert{} model introduced in~\cite{gunther2023jina2}.
To reduce the number of parameters, we use a different tokenizer in each bilingual model that is trained to encode the specific pair of languages.
As with \JBert{}, those bilingual models are able to encode up to 8192 tokens.
For training the model from scratch, we conduct a standard masked language modeling training with a combination of large full-text corpora in both languages.\label{stage:one}
\item \textbf{Fine-tuning with Text Pairs:} To encode a text passage into a single vector representation, we employ the previously used method~\cite{gunther2023jina2}, which fine-tunes a model on text pairs.\label{stage:two}
\item \textbf{Fine-tuning with Multi-Task Objective:} In the last stage, training with a multi-task objective is done to specifically improve the model's performance on STS, retrieval, and classification tasks.\label{stage:three}
\end{enumerate}

% While both stages~\ref{stage:two} and~\ref{stage:three} are geared towards training the models for general embedding tasks, the latter is especially critical for improving the model's performance in retrieval and classification tasks (refer to Section~\ref{sec:evaluation_jina_embedding}).

\section{Pre-training a Modified BERT}
\label{sec:backbone_pretraining}

\subsection{Model Architecture}
\label{sec:model_architecture}

For our bilingual text embedding models, we use the same BERT architecture described in our previous work~\cite{gunther2023jina2}. This architecture modifies the original BERT model with ALiBi~\cite{press2022alibi} to support long context lengths during model inference. 
For the Spanish model, we additionally introduce a normalization of key and query vectors to address training instabilities. In each self-attention layer, we use layer normalization~\cite{ba2016layer} to normalize keys and queries before computing the dot-product attention. For the purpose of normalization, we decide to treat the head axis as an additional layer axis.
This normalization is loosely related to QK-normalization~\cite{henry2020query}, in which keys and queries are $\ell^2$-normalized. In contrast to~\cite{henry2020query}, we do not perform normalization in each head separately, but instead flatten the head axis into the layer axis. 

Another modification to the model architecture has to do with the tokenizer and the vocabulary size. We train BPE tokenizers ~\cite{sennrich2016neural} for all the language pairs and double the vocabulary size compared to the original monolingual variant to provide room for the token representation of two languages instead of one. This increase in the vocabulary results in a larger input embedding matrix, thus increasing the number of parameters for our bilingual models, as is apparent in Table~\ref{tab:parameter-count}.

\begin{table}
\small{
\begin{tabular}{ lrrrr  }
 \toprule
 Model & Layers & Dim. & Param. & Tokens \\
 \midrule
% \JBert~Small & 4 & 512 & 33M \\
%\multirow{ 2}{*}{\shortstack[l]{BERT-base- \\ uncased (en)}} & \multirow{ 2}{*}{12} & \multirow{ 2}{*}{768} & \multirow{ 2}{*}{110M} & \multirow{ 2}{*}{30,522} \\\\\midrule
\multirow{ 2}{*}{\shortstack[l]{\JBert \\ Base (en)}} & \multirow{ 2}{*}{12} & \multirow{ 2}{*}{768} & \multirow{ 2}{*}{137M} & \multirow{ 2}{*}{30,528}\\\\
\multirow{ 2}{*}{\shortstack[l]{\JBert \\ Base (bilingual)}} & \multirow{ 2}{*}{12} & \multirow{ 2}{*}{768} & \multirow{ 2}{*}{161M} & \multirow{ 2}{*}{\shortstack[l]{[60,516, \\61,056]}}\\\\
\multirow{ 2}{*}{\shortstack[l]{multilingual- \\ e5-base}} & \multirow{ 2}{*}{12} & \multirow{ 2}{*}{768} & \multirow{ 2}{*}{278M} & \multirow{ 2}{*}{250,000}\\\\
% JBert~Large & 24 & 1024 & 455M \\
 \bottomrule
\end{tabular}
}
\caption{Architecture specifications of different monolingual, bilingual and multilingual models}\label{tab:parameter-count}
\end{table}

\subsection{Training Data}
\label{sec:dataset:full-text}

We collect a comprehensive set of data from various high-quality sources like CulturaX~\cite{nguyen2023culturax}, Wikipedia\footnote{\url{https://dumps.wikimedia.org} (Accessed: 2024-02-20)}, and Opus~\cite{tiedemann-2016-opus}. CulturaX is a combination of mC4~\cite{2019t5}~(version 3.1.0) and OSCAR~\cite{oscar} aggregating all accessible corpora up to 23.01.2024. The Opus collection consists of translated texts harvested from the Web and is publicly available. To ensure the high quality of the dataset, the original creator of the data source conducted a thorough cleaning process, wherein various filtering rules and heuristics were applied along with multiple stages of deduplication. We amass a corpus of approximately 250M English text documents and an equivalent number of German and Spanish text documents for each of the respective bilingual models, amounting to a total of 256B tokens for each model. This meticulous curation maintains a 1:1 ratio between English and target language text documents, ensuring balance and relevance for their respective bilingual models in our study. We earmark 1\% of the dataset exclusively for the assessment of the model performance during training, i.e. calculating the validation loss and measuring accuracy in masked token prediction.

% For the pre-training phase, we leverage the English ``Colossal, Cleaned, Common Crawl (C4)'' dataset~\footnote{\url{https://huggingface.co/datasets/c4}}, encompassing approximately 365 million text documents harvested from the web, summing to around 170 billion tokens. As delineated in ~\cite{raffel2020exploring}, the C4 dataset is a refined iteration of Common Crawl, utilizing heuristics for cleanup and language recognition, retaining solely English content. As a result, our models are monolingual and tailored exclusively for English texts. The purification process also encompasses the removal of webpages hosting inappropriate content. We reserve $1\%$ of the dataset for evaluating validation loss and the accuracy of the masked language modeling (MLM) task.

\subsection{Training Algorithm}
\label{sec:training_algorithm}

The training algorithm largely remains the same as in our previous work~\cite{gunther2023jina2}. We optimize for the Masked Language Modeling (MLM) objective, omitting the Next Sentence Prediction (NSP) task ~\cite{liu2019roberta}. We employ whole word masking with a masking rate of $30\%$. Experiments with dynamic masking rate scheduling ~\cite{ankner2024dynamic} did not show any benefits on downstream task performance for our use case.

%The output embedding matrix, used by the language modeling head to resolve the masked tokens, is tied to the input embedding matrix. Using different matrices (and dimensionalities) for the input and output embeddings did not bring significant improvement in our experiments.    

% Given the fact that our training dataset is bilingual, we experiment with two different strategies for consuming the data during training.
% In the first option, we present data from both languages at every batch with a 1:1 ratio, while in the second option, we alternate between batches containing data from a single language.
% Both batching strategies show comparable performance on downstream tasks and convergence speed. 
% Eventually, we use the latter monolingual batch strategy due to the training loss curve appearing smoother in the monolingual batching experiments. 
Given the fact that our training dataset is bilingual, we alternate between batches containing data from a single language.

All other parameters and mechanics of our training algorithm, e.g., optimizer, learning rate, schedulers, and hyperparameters are identical to our monolingual model and are presented extensively in our previous work~\cite{gunther2023jina2}.

\section{Fine-Tuning for the Embedding Task}
\label{sec:embedding-training}
% @michael and @george and @isabelle
After pre-training the \JBert{} models, we train the models to encode a text sequence into a single vector representation.
Following the Sentence-BERT~\cite{reimers2019sentence} approach, each model is augmented by a mean pooling layer to consolidate the semantic encoded in all the output token vectors into a single vector.
Therefore, we conduct generic embedding training in stage~\ref{stage:two}, which is followed by stage~\ref{stage:three}, where the model learns to solve specific embedding tasks.

To increase the run-time performance and reduce the consumption of GPU memory to support larger batch sizes (high batch sizes are crucial for effective embedding training~\cite{li2023general}), we apply various optimizations explained in detail in~\cite{gunther2023jina2}.

\subsection{Fine-tuning with Text Pairs}
\label{subsec:unsupervised_pretraining}
In stage~\ref{stage:two}, we train the models for general embedding tasks on a corpus of text pairs $(q,p)$, comprising a query string $q$ and a target string $p$.

\paragraph{Training Data}
We construct a collection of 211 million German, 111 million Spanish, and 518 million English text pairs. The dataset predominantly consists of monolingual pairs (97.5\%), which include varied relational types like titles with articles, questions with answers, and texts with their summaries. 
The rest are bilingual pairs that consist of identical texts in two different languages, forming a parallel dataset that bridges English with the other two.

To assemble our dataset, we start with curated datasets such as MQA~\cite{de2021mfaq}, XLSum~\cite{hasan-etal-2021-xlsum}, XNLI~\cite{conneau2018xnli}, MLSum~\cite{scialom2020mlsum}, and Europarl~\cite{tiedemann2012parallel}. Seeking further diversity and volume, we utilize Common Crawl data to create two types of pairs: one from web page titles and their contents, and another by mining question-answer pairs from FAQ and support-related pages.
Additionally, we pair paragraphs with their section titles from Wikipedia.

We improve the data quality by implementing a two-stage processing strategy: filtering and refinement. In the filtering stage, texts are eliminated based on a variety of quality checks, e.g., very short texts, lengthy texts, and text with excessive repetition of lines or n-grams are removed~\cite{rae2021scaling}. Then the refinement process is applied to improve the remaining texts' quality without discarding them. This includes stripping site-specific metadata, and removing unfinished lines, or lines with only one word.

To further improve our dataset, we use near-deduplication~\cite{lee2022deduplicating} to remove duplicates and consistency filtering~\cite{alberti2019synthetic, dai2022promptagator} to filter out inconsistent pairs.
These techniques have proven effective in our past studies~\cite{gunther2023jina, gunther2023jina2}.
\paragraph{Loss Function}
Following our previous work, we utilize a bi-directional InfoNCE~\citep{DBLP:journals/corr/abs-1807-03748} function $\mathcal{L}_{\mathrm{pairs}}$ with a temperature value of $\tau=0.05$:
\iffalse
\begin{flalign}
    \mathcal{L}^{\mathrm{pairs}}(B) &:= \mathcal{L}^{\mathrm{pairs}}_{\nce}(B) + \mathcal{L}^{\mathrm{pairs}}_{\overline{\nce}}(B),\text{ with} \nonumber\\
    \mathcal{L}_{\overline{\nce}}^{\mathrm{pairs}}(B) &:= \mathbb{E}_{(q,p)\sim B}\left[-\ln \frac{e^{s(p, q) / \tau}}{\sum\limits_{i = 1}^k e^{s(p, q_i) / \tau}}\right]
\end{flalign}
\fi
\begin{equation}
    \mathcal{L}_{\mathrm{pairs}}(B) := \mathcal{L}_{\nce}(B) + \mathcal{L}_{\nce}(B^\dagger)
\end{equation}
where $B = ((p_1, q_1),\ldots,(p_k, q_k))$ is a batch of $k$ pairs, $B^\dagger= ((q_1, p_1),\ldots,(q_k, p_k))$ is obtained from $B$ by swapping the order of pairs and $\mathcal{L}_{\nce}$ is defined via:
\begin{equation}
    \mathcal{L}_{\nce}(B) := -\sum_{(x_i, y_i) \in B} \ln \frac{e^{s(x_i, y_i) / \tau}}{\sum\limits_{i' = 1}^k e^{s(x_i, y_{i'}) / \tau}}
\end{equation}

% The constant temperature parameter $\tau$ influences how the loss function weighs minor differences in the similarity scores~\cite{wang2021understanding}. Empirical testing suggests that $\tau=0.05$ is effective.

%
\subsection{Fine-tuning with Multi-Task Objective}
\label{sec:supervised_fine_tuning}
% @michael and @bo

Previous work~\cite{wang2022text, li2023general, gunther2023jina2} found that models trained with the training paradigm in stage~\ref{stage:one} and stage~\ref{stage:two} have sub-optimal ranking capabilities that can be improved by continuing training with hard negatives and different contrastive loss functions.
We generalize from this concept to a multi-task training stage.

For different tasks, datasets come with different formats.
For example, STS tasks~\cite{cer2017semeval} usually contain triplets $(q, p, t)$ of two texts $q$ and $p$ associated with their defined similarity score $t$ while retrieval datasets usually contain a text query $q$ and one or several relevant documents $p$ and optionally non-relevant documents $n$.
For each task, we define a specific loss function.
For this work, we only focus on retrieval and STS tasks, but plan to include more tasks in our training in the future.
Similar to the training in stage~\ref{stage:two}, for the construction of each batch, a dataset is sampled.
Depending on the task this dataset aims to solve, we choose the respective loss function.

% For retrieval tasks, datasets with hard negatives in the same format as previous work~\cite{gunther2023jina2} are used while for sentence similarity tasks datasets with similarity scores are used instead.
% \todo{continue description ...}

% The goal of the supervised fine-tuning stage is to improve the models' ranking capabilities. This improvement is achieved by training with datasets that include additional negative examples.

\paragraph{Training Data} 

For retrieval tasks, we prepare several datasets like MSMarco~\cite{bajaj2016ms} and Natural Questions (NQ)~\cite{47761} into a format of tuples with one positive and multiple negatives: $(q, p, n_1, \ldots, n_{m})$ similar to previous work~\cite{li2023general,gunther2023jina2}.
The STS datasets contain triplets of two text values $q$ and $p$ as well as a similarity score $t$: $(q, p, t)$ like the training datasets of SemEval\footnote{\url{https://ixa2.si.ehu.eus/stswiki/index.php/STSbenchmark}}.

Since there is much more high-quality data available in English, we use machine translation to translate some of the datasets into the other target languages. 
For translating the datasets to German, we employ FAIR's compute-efficient WMT19 model~\citep{ng2019wmt19}, whereas we utilize the multilingual MADLAD-3B model of~\citet{kudugunta2023madlad} for generating Spanish datasets.
In both cases, translations are performed at the sentence level. 
Translated documents containing truncations, either originating from tokenization of the original text or generation, are removed from the dataset.

To increase the diversity of the training data and mitigate problems caused by training only on translations, we also select various datasets in the target language that we prepare for the final training stage.
We make use of GerDaLIR~\cite{wrzalik-krechel-2021-gerdalir}, which is a German Dataset for legal information retrieval, as well as GermanDPR~\cite{möller2021germanquad}, which is a German training dataset for dense passage retrieval. 
%We also prepare test split of the GermanDPR dataset as a retrieval task in the MTEB, with further details to be found in Section~\ref{ssec:embedding-eval}.

For Spanish, we use the SQAC dataset \cite{maria-esqac}, which is an extractive QA dataset for Spanish, as well as the Spanish subset of XL-Sum~\cite{hasan-etal-2021-xlsum}, which is a multilingual summarization dataset consisting of BBC news articles. 
We also use the Spanish subset of the MIRACL dataset~\cite{zhang2022miracl}. This is a multilingual information retrieval dataset, which we also use to prepare new multilingual retrieval and re-ranking tasks in the MTEB. See Section~\ref{sec:embedding-eval} for more details.

% We have prepared retrieval datasets, such as MSMarco~\cite{bajaj2016ms} and Natural Questions (NQ)~\cite{47761}, in addition to multiple non-retrieval datasets like the Natural Language Inference (NLI) dataset~\cite{bowman2015large}. These datasets encompass a collection of queries with annotated relevant passages and several negative examples, consistent with earlier work~\cite{wang2022text}. Each training batch $B$, structured as $(q, p, n_1, \ldots, n_{15})$, includes one positive and 15 negative instances. For retrieval datasets, hard negatives are discerned by identifying passages deemed similar by retrieval models. This approach instructs the model to prioritize relevant documents over those that are merely semantically related. For non-retrieval datasets, negatives are selected randomly, since drawing a clear line between positives and hard negatives isn't feasible. This is because, unlike relevancy, it's challenging to make a binary determination regarding the similarity or dissimilarity of two textual values. Consequently, opting for hard negatives in such datasets seemed to diminish the models' quality. Nonetheless, it remains crucial to integrate these datasets into the stage~\ref{stage:three} training to ensure continued performance on non-retrieval tasks. To ensure that hard negative passages are indeed less relevant than the annotated relevant ones, we employ a cross-encoder model to validate that their relevance score is indeed lower.

\paragraph{Loss Function}
For each batch, the type of task is determined and a respective loss function is selected.
For retrieval tasks, we follow previous work and use a bi-directional InfoNCE loss function:
% Our training employs a modified variant of the InfoNCE loss function, denoted as $\mathcal{L}_{\nce^+}$ and described by Equation \eqref{eq:loss-hard-negatives}. Similar to the preceding loss function, this one is bidirectional and incorporates the additional negatives when pairing queries with passages:
\begin{flalign}
&\mathcal{L}_{\text{triplet}}(B) := \nonumber\\
&\;\;\;\;\;\mathbb{E}_{r\sim B}\Bigg[-\ln \frac{e^{s(q, p) / \tau}}{\sum\limits_{i = 1}^k \Big[ e^{s(q, p_i) / \tau}+ \sum\limits_{j = 1}^{m} e^{s(q, n_{j,i}) / \tau}\Big]}\Bigg]\nonumber \\
&\, + \mathbb{E}_{r\sim B}\Bigg[-\ln \frac{e^{s(p, q) / \tau}}{\sum\limits_{i = 1}^k e^{s(p, q_i) / \tau}}\Bigg]\nonumber \\
&\text{with}\; r = (q,p, n_1, \ldots, n_{m}).\label{eq:loss-hard-negatives}
\end{flalign}
If the sampled batch comes from an STS dataset, we employ the negative Pearson's sample correlation~\citep{1895pearson} as a loss function. Assuming that the batch $B$ consists of text pairs $(q, p)$ and corresponding relevance scores $t$, we compute it as:
\begin{equation}
    \mathcal{L}_{\text{STS}}(B) := -\frac{\mathrm{cov}_{(q, p, t)\sim B}(s(q, p), t)}{\sigma_s(B) \sigma_t(B)}
    \label{eq:pearson_correlation}
\end{equation}
Here, $\sigma_s$ and $\sigma_t$ are the estimated standard deviations of $s(q, p)$ and $t$, respectively, computed over batch $B$. Analogously, the covariance in the numerator of~\eqref{eq:pearson_correlation} is computed across batch $B$.

The correlation-based loss is suitable because it takes into account the magnitudes of the similarity scores instead of binary relevance as considered by the InfoNCE.
Moreover, we choose to use a Pearson correlation loss over the mean squared error (MSE) loss function which is typically used with STS datasets~\cite{reimers2019sentence} since the Pearson correlation is invariant to the scale of the similarity values.
We presume that this makes multi-task training that involves the InfoNCE loss function for retrieval datasets more stable.
The ablation study in Section~\ref{sec:sts-ablation} provides empirical evidence for this claim.
%
%Moreover, it is better compatible with the InfoNCE loss than mean squared error (MSE) loss functions which are typically used with STS datasets~\cite{reimers2019sentence} since it provides more freedom regarding the scale of the similarity values, i.e., a regression loss function like MSE would impose the model to assign exactly the similarity values defined in the dataset.
%Empirical evidence for this calim is provided in ablation study in Section~\ref{sec:sts-ablation}
%\todo{link evaluation section if we have an ablation study}
%
\section{Evaluation}
\begin{table*}[ht]
    \small
    \centering
    \setlength\tabcolsep{2.5pt}
    \begin{tabular}{ lrcccccccccc  }
 \toprule
 Model& Params & MNLI & QQP & QNLI & SST-2 & CoLa & STS-B & MRPC & RTE & WNLI & Average \\
 \midrule
\JBert~Base-de & 161M & 84.4/85.1 & 89.8 & \bfseries 91.3 & 92.6 & 57.9 & 89.1 & 87.5 & 79.6 & \bfseries 56.3 & 81.0 \\
\JBert~Base-es & 161M & \bfseries 86.3/86.2 & \bfseries 90.2 & 91.2 & \bfseries 93.6 & \bfseries 61.7  & \bfseries 90.1 & \bfseries 89.2 & \bfseries 82.3 & 53.3 & \bfseries 82.0 \\
XLM-RoBERTa & 279M & 84.7/84.7 & 90.1 & 90.6 & 93.1 & 56.9 & 88.7 & 88.2 & 76.2 & \bfseries 56.3 & 80.8 \\
mBERT Base & 179M & 82.5/82.4 & 89.5 & \bfseries 91.3 & 90.1 & 42.6 & 89.9 &  87.0 & 79.0 & 54.9 & 78.2 \\
\bottomrule
\end{tabular}
    \caption{Performance of \JBert{} models and multilingual models on GLUE dev set}
    \label{table:glue-performance}
\end{table*}
\begin{table*}[ht]
    \small
    \centering
    \setlength\tabcolsep{2.5pt}
    \begin{tabular}{llcccccccc}
 \toprule
 Lang. & Model & XNLI & PAWS-X & NER & XQuAD & MLQA & BUCC & Tatoeba\\
\midrule
\multicolumn{2}{l}{Metrics} & Acc. & Acc. & F1 & F1/EM & F1/EM & F1 & Acc. \\
\midrule
\multicolumn{10}{l}{\textit{Cross-lingual zero-shot transfer (models are trained on English data)}} \\
\midrule
\multirow{3}{*}{de}  & \JBert~Base-de & \bfseries 79.8 & 85.8 &  77.3 & 75.4/58.7  &\bfseries  62.5/46.3 & \bfseries 72.5  & \bfseries 86.2  \\
& XLM-RoBERTa   & 77.6 & \bfseries 87.7 &  75.8 & \bfseries 75.5/59.9  &  61.9/45.7 & 17.9  &  51.6  \\
& mBERT         &  72.8 & 84.6 &  \bfseries 79.3 & 73.2/56.9  &  56.7/39.2 & 65.8  &  60.9  \\
\midrule
\multirow{3}{*}{es} & \JBert~Base-es  & \bfseries 82.6 &  89.6 &  73.2 & \bfseries 78.2/60.1 & \bfseries 72.7/50.2 & -  & \bfseries 76.6 \\
& XLM-RoBERTa   & 78.9 &  \bfseries 89.8 &  72.0 & 77.0/58.5  &  70.3/49.2 & -  &  32.2   \\
& mBERT         & 75.5 &  87.3 &  \bfseries 76.3 & 71.2/53.5  &  67.6/46.8 & -  &  57.1  \\
\midrule
\multicolumn{10}{l}{\textit{Translate-train (models are trained on English training data translated to the target language)}} \\
\midrule
\multirow{3}{*}{de}  & \JBert~Base-de & \bfseries 81.4 & \bfseries 87.4 &  88.7 & 75.6/59.8  & \bfseries63.0/45.5 & -  &  -  \\
&XLM-RoBERTa    & 80.6 &  86.6 &  88.9 & 76.7/60.3  &  62.2/47.1 & -  &  -  \\
&mBERT          & 76.9 &  85.4 & \bfseries 90.4 & \bfseries76.3/60.8  &  59.9/44.4 & -  &  -   \\
\midrule
\multirow{3}{*}{es} & \JBert~Base-es  &\bfseries 84.2 & \bfseries 90.3 &  90.3 & 80.0/61.5 & 72.8/49.2 & -  &  -  \\
&XLM-RoBERTa    & 80.9 &  89.0 &  90.5 & 79.6/62.6  &  71.1/48.6 & -  &  -  \\
&mBERT          &  79.4 &  88.8 &  \bfseries91.9 &\bfseries 80.6/62.8  & \bfseries 72.6/51.2 & -  &  - \\
 \bottomrule
\end{tabular} % Jina models have 161M parameters. We mark entries that are the average of just one value with a * 
    \caption{Performance of \JBert{} models and multilingual models on XTREME dev set}
    \label{table:xtreme}
\end{table*}
In our evaluation, we aim to not only evaluate the resulting models on a wide range of tasks, but also evaluate our novel approach.
We start evaluating the bilingual backbone models on general language modeling tasks in Section~\ref{sec:eval:backbone_performance}.
Then, we conduct a comprehensive evaluation concerning embedding tasks in Section~\ref{sec:embedding-eval} and finally conduct two ablation studies to show that bilingual models perform better after training on embedding tasks with the same amount of pair-wise training data (Section~\ref{sec:ablation_bilingual}) and prove the effectiveness of the multi-task learning (Section~\ref{sec:sts-ablation}).
\label{sec:evaluation}
\begin{table*}[htb]
\
    \centering
    \setlength{\tabcolsep}{4.5pt} 
    \small{
        \begin{tabular}{llccccccccc}
 \toprule
 Lang. & Model & CF & CL & PC & RR & RT & STS** & SM \\
 \midrule
    \multirow{3}{*}{en}  &\JinaDe & 0.688 &  0.403 &  0.835 & 0.549  &  0.441 & 0.820  & \bfseries 0.318*    \\
    &\JinaEs & 0.690  &  \bfseries 0.405 & \bfseries 0.846  & \bfseries 0.553  & 0.464 &  \bfseries 0.835 & 0.299*    \\
    %&\JinaZh &   0.677 &  0.368 &  0.815 &  0.550 & 0.460   & 0.826  & 0.313*  \\
    &\MultilingualEFiveBase & \bfseries 0.730 & 0.400 & 0.836  & 0.548 & \bfseries 0.489 & 0.803 & 0.301*   \\
    \midrule
    \multirow{2}{*}{de} & \JinaDe  &  0.665 &  0.299 &  \bfseries 0.583* & 0.639  & \bfseries 0.387  &  \bfseries 0.714 &  --  \\
    & \MultilingualEFiveBase &  \bfseries 0.687 & \bfseries 0.328 &  0.541* &  \bfseries 0.648 &  0.342 &  0.674 &  --   \\
    \midrule
    \multirow{2}{*}{es} & \JinaEs &  0.671 & \bfseries 0.440  &  \bfseries 0.583* & \bfseries 0.739  &  \bfseries 0.511 & \bfseries 0.788  &  --   \\
    & \MultilingualEFiveBase & \bfseries 0.685  & 0.413  & 0.541*  &  0.736 &  0.478 &   0.783 &  -- &   \\
%jina-embeddings-large-v2 & 435M &  &  &  &  &  &  &  &  \\
 \bottomrule
\end{tabular} % Jina models have 161M parameters. We mark entries that are the average of just one value with a * 

%\bfseries 0.583 & 0.541 \\
        
        CF: Classification Accuracy \quad{}
        CL: Clustering $\mathcal{V}$ measure\quad{}
        PC: Pair Classification Average Precision\quad{} \\
        RR: Reranking MAP\quad{}
        RT: Retrieval nDCG@10\quad{}
        STS: Sentence Similarity Spearman Correlation\quad{} \\
        SM: Summarization Spearman Correlation\quad{} \\
        * Values are the average over a single task.\quad{}
        ** Includes an MTEB datasets (STSBenchmark) that evaluates partly on train data.
    }
    \caption{Evaluation of the \texttt{Jina}~models on MTEB tasks. We consider the subsets of English (en), German (de), and Spanish (es) tasks separately. }
    \label{tab:mteb_results}
\end{table*}
\subsection{Performance of Backbone Models}
\label{sec:eval:backbone_performance}
To assess the proficiency of our bilingual backbone models, we conduct evaluations on two distinct benchmarks for language models.
The GLUE (General Language Understanding Evaluation)~\cite{wang2019glue} benchmark is used to examine the model's capabilities in the English language.
It comprises nine datasets designed to evaluate natural language understanding specifically in English.
For measuring the cross-lingual performance of our model, we utilize the XTREME (Cross-lingual TRansfer Evaluation of Multilingual Encoders) benchmark~\cite{hu2020xtreme}.
XTREME is a comprehensive multi-task benchmark designed to evaluate the cross-lingual generalization capabilities of multilingual representations.
For both benchmarks, the training, development, and test data splits are provided.
It is worth noting that the test splits do not include labels, and the participants can evaluate their systems by submitting predictions on the test set via the designated submission servers.\footnote{\url{https://gluebenchmark.com}}$^{,}$\footnote{\url{https://sites.research.google/xtreme}}
In our study, we conduct a comparative analysis of our models against established multilingual models, specifically mBERT~\cite{devlin2019bert} and XLM-RoBERTa~\cite{conneau2020xlmr}, which are commonly used as backbones for multilingual embedding models.
The evaluation is performed on the development sets of the two benchmarks.

For evaluation on the GLUE benchmark, we follow the methodology of \cite{phang2018mnlitrick},  for RTE, STS, and MRPC, and start fine-tuning from models trained on GLUE's MNLI task, rather than utilizing the pre-trained models directly.

For the XTREME benchmark, our experiments involve two distinct settings.
Firstly, we concentrate on zero-shot cross-lingual transfer, employing English as the source language and German and Spanish as the target languages.
In the second setting, translated train splits are available to conduct an evaluation where we train our models on the German and Spanish translations.

In both benchmarks, we train for 10 epochs with a batch size of $32$ and a learning rate of $2\mathrm{e}{-5}$.
For each task, we report the best performance on the development set.

Table \ref{table:glue-performance}, and Table~\ref{table:xtreme} report the results of the GLUE and XTREME benchmarks.
Across both benchmarks our bilingual models perform on average better than the multilingual ones.
Notably, our Spanish model exhibits the highest performance on the GLUE benchmark, while the German model slightly surpasses the XLM-RoBERTa model.
On the XTREME benchmark, our bilingual models exhibit a noteworthy advantage in the zero-shot cross-lingual evaluation setup.
Particularly, they outperform multilingual models in cross-lingual sentence retrieval tasks of BUCC~\cite{zweigenbaum2017overview} and Tatoeba~\cite{artetxe2019massively}.
This difference margin decreases after training on the translated data.
\subsection{Embedding Model Evaluation}
\label{sec:embedding-eval}
To get a comprehensive overview of our models on a broad range of tasks, we evaluate them on the MTEB~\cite{muennighoff2023mteb} which consists of more than 50 embedding evaluation tasks.
Although the MTEB boasts a large variety of tasks and languages, not all languages are adequately represented across the existing task categories, such as retrieval, classification, and reranking.
The lack of tasks for German and Spanish hinders a comprehensive evaluation of our bilingual models trained on these languages.
Consequently, we address these gaps by integrating new tasks into the MTEB framework, using existing publicly available datasets for the aforementioned languages, to ensure a more inclusive and representative assessment of model performance across diverse languages and cross-lingual combinations.
Using the MTEB framework, others can easily reproduce the results and evaluate other new and existing embedding models.

We add a total of ten tasks for Spanish, which are comprised of six retrieval tasks (XMarket~\cite{bonab2021crossmarket}, SpanishPassageRetievalS2P and SpanishPassageRetrievalS2S~\cite{kamateri2019spanishpassageretrieval}, MIRACLRetrieval~\cite{zhang2022miracl}, MintakaRetrieval~\cite{sen-etal-2022-mintaka}, and XQPARetrieval~\cite{shen2023xpqa}), two clustering tasks (FloresClusteringS2S~\citet{goyal2021flores} and SpanishNewsClusteringP2P\footnote{https://www.kaggle.com/datasets/kevinmorgado/spanish-news-classification}), as well as one reranking task (MIRACLReranking~\cite{zhang2022miracl}) and one STS task (STSES~\citep{agirre2014semeval, agirre2015semeval}. 
Additionally, we integrate the PAWS-X dataset \cite{pawsx2019emnlp} as a multilingual pair-classification task for seven different languages, including German and Spanish.

Table~\ref{tab:mteb_results} shows the averages across task categories per model.
Our Spanish model outperforms multilingual-e5 on average on the Spanish clustering, retrieval, and reranking tasks, while slightly underperforming in the classification domain. However, our bilingual Spanish model also has an advantage over multilingual-e5 on cross-lingual STS tasks (see Table~\ref{table:spanish_english_mteb}).
Details on the evaluation results of our Spanish bilingual model on all Spanish tasks in the MTEB are presented in Table~\ref{table:spanish_mteb} in the appendix.

We also integrate five new tasks for German, namely a reranking task (MIRACLReranking~\cite{zhang2022miracl}), an STS task (GermanSTSBenchmark\footnote{https://github.com/t-systems-on-site-services-gmbh/german-STSbenchmark}), and three retrieval tasks (GerDaLIR~\cite{wrzalik-krechel-2021-gerdalir}, GermanDPR~\cite{möller2021germanquad} and XMarket~\cite{bonab2021crossmarket}).

Looking to Table~\ref{tab:mteb_results} for average scores across task categories, our bilingual German model outperforms the multilingual-e5 model on the German retrieval tasks, as well as on the STS tasks, and achieves relatively comparable results on the classification, clustering, and reranking tasks. 
The full results for these tasks can be found in Table~\ref{table:german_mteb}, together with all other German tasks in the MTEB. 
Table~\ref{table:german_english_mteb} reports the performance on cross-lingual German-English tasks, showing that our German bilingual model either matches or outperforms that of multilingual-e5.

In our efforts to enhance our model with bilingual functionalities without sacrificing its efficiency in monolingual English tasks, it is important to observe the performance of our German and Spanish bilingual models across all English tasks, as detailed in Table~\ref{table:english_mteb}.
Consistency in average scores across task domains is maintained among the diverse bilingual models, with the bilingual Spanish model securing the highest average scores in various tasks.
It is only in the domains of classification and retrieval tasks that the multilingual-e5 base model outperforms others on average.
This indicates a balanced expansion of bilingual capabilities in our model, ensuring enhanced performance without compromising its efficiency in English-specific tasks.

\subsection{Comparing Bilingual to Multilingual Models}
\label{sec:ablation_bilingual}
\begin{table}[ht!]
    \small
    \centering
    \setlength\tabcolsep{1.5pt}
    \begin{tabular}{lcccc}
\toprule
Datasets & JinaBert & BiBERT & mBERT & XLM-R \\
\midrule
\multicolumn{5}{l}{\bfseries German Tasks} \\ % DE Tasks: & 0.571 & \bfseries 0.601 & 0.553 & 0.551 \\
Ger.DPR\textsuperscript{RT} & 0.738 & \bfseries 0.772 & 0.760 & 0.719 \\
Ger.STSBenchmark\textsuperscript{STS} & 0.768 & \bfseries 0.788 & 0.716 & 0.751 \\
MIRACL\textsuperscript{RR} & 0.600 & \bfseries 0.615 & 0.593 & 0.570 \\
PawsX\textsuperscript{PC} & 0.598 & \bfseries 0.606 & 0.604 & 0.599 \\
STS22\textsuperscript{STS} & 0.527 & \bfseries 0.574 & 0.476 & 0.498 \\
XMarket\textsuperscript{RT} & 0.193 & \bfseries 0.251 & 0.172 & 0.167 \\
 DE Average & 0.571 & \bfseries 0.601 & 0.553 & 0.551 \\
\midrule
\multicolumn{5}{l}{\bfseries Cross-lingual Tasks} \\ % \bfseries Cross-lingual Tasks: & \bfseries 0.679 & 0.673 & 0.617 & 0.633 \\
STS22 (de-en)\textsuperscript{STS} & \bfseries 0.575 & 0.554 & 0.530 & 0.536 \\
STS17 (en-de)\textsuperscript{STS} & 0.784 & \bfseries 0.793 & 0.704 & 0.731 \\
CL Average & \bfseries 0.679 & 0.673 & 0.617 & 0.633 \\
\midrule
\multicolumn{5}{l}{\bfseries EN Tasks} \\ % \bfseries EN Tasks: & 0.759 & \bfseries 0.764 & 0.748 & 0.751 \\
QuoraRetrieval\textsuperscript{RT} & 0.850 & \bfseries 0.863 & 0.832 & 0.846 \\
STS12\textsuperscript{STS} & 0.703 & 0.714 & \bfseries 0.725 & 0.714 \\
STS13\textsuperscript{STS} & 0.793 & \bfseries 0.834 & 0.793 & 0.811 \\
STS14\textsuperscript{STS} & 0.736 & \bfseries 0.760 & 0.737 & 0.738 \\
STS15\textsuperscript{STS} & 0.831 & \bfseries 0.842 & 0.832 & 0.839 \\
STS16\textsuperscript{STS} & 0.796 & \bfseries 0.805 & 0.785 & 0.792 \\
STS17\textsuperscript{STS} & \bfseries 0.860 & 0.857 & 0.842 & 0.833 \\
STS22\textsuperscript{STS} & \bfseries 0.675 & 0.671 & 0.661 & 0.663 \\
TRECCOVID\textsuperscript{RT} & \bfseries 0.585 & 0.527 & 0.529 & 0.525 \\
EN Average & 0.759 & \bfseries 0.764 & 0.748 & 0.751 \\
\bottomrule
\end{tabular}
    \label{table:mlm-ablation-study}
    \caption{Evaluation of multilingual and bilingual models after short embedding training on pairs}
    \label{tab:ablation_mlm}
\end{table}
\begin{table*}[ht]
    \centering
    \small
    \begin{tabular}{llcccccccccc}
     \toprule
     Lang. & Variant & BIOSSES & SICK-R* & STS12* & STS13 & STS14 & STS15 & STS16 & STS17 & STS22  \\
     \midrule
     \multirow{3}{*}{de} 
      & Pearson  & 0.780 &  \bfseries 0.811 & \bfseries 0.801 & \bfseries 0.851 & 0.811 & \bfseries 0.878 & 0.836 & \bfseries 0.893 & \bfseries 0.682 \\
      & MSE & 0.784 & 0.793 & 0.797 & 0.849 & \bfseries 0.812 & 0.874 & \bfseries 0.842 & 0.892 & 0.677 \\
      & No STS data & \bfseries 0.797 & 0.791 & 0.792 & 0.847 & 0.810 & 0.876 & 0.839  & 0.890 & 0.676 \\
     \midrule
     \multirow{ 3}{*}{es} 
     & Pearson & \bfseries 0.830 & \bfseries 0.830 & \bfseries 0.816 & \bfseries 0.868 & \bfseries 0.836 & \bfseries 0.887 & \bfseries 0.860 & \bfseries 0.886 & 0.656 \\
     & MSE & 0.825 & 0.801 & 0.810 & 0.867 & 0.834 & \bfseries 0.887 & \bfseries 0.860 & 0.884 &     0.657 \\
     & No STS data & 0.825 &  0.792 & 0.801 & 0.866 & 0.831 & 0.882 & 0.855 & 0.882 & \bfseries 0.662 \\ 
     \bottomrule
\end{tabular}
    \begin{minipage}{0.97\textwidth}\raggedright
    \vspace{2mm}
    * Corresponding training datasets used during the training
    \end{minipage}
    \caption{Spearman correlation on English STS tasks after triplet-tuning}
    \label{tab:ablation_study_sts}
\end{table*}
One motivation for training bilingual models instead of multilingual embedding models is the assumption that bilingual models achieve greater performance in both languages.
To obtain empirical evidence for this claim, we evaluate two German-English bilingual and two multilingual transformer models on embedding tasks after fine-tuning with a small set of German and English text pairs.
In particular, we use our German-English model, the bilingual BiBERT model~\cite{Xu2021bibert} as well as the multilingual mBERT~\cite{devlin2019bert} and XLM-RoBERTa~\cite{conneau2020xlmr} models.
We use the same stage~\ref{stage:two} training setup as described in Section~\ref{subsec:unsupervised_pretraining}, however, we restrict the training to $1000$ steps and a single GPU node which involves roughly 2M pairs in the training.
We omit multi-task learning from this training.
The results presented in Table~\ref{tab:ablation_mlm} show that the bilingual models perform clearly better on the German and cross-lingual tasks and also slightly outperform their multilingual counterparts on the English tasks.
\subsection{Performance of Multi-Task Learning}
\label{sec:sts-ablation}
To investigate the effectiveness of our multi-task learning strategy (see Section~\ref{sec:supervised_fine_tuning}), we analyze the contribution of including STS training data with an additional loss function.
We study three fine-tuned model variants of the Spanish and German models obtained from the stage~\ref{stage:two} training.
The first variant (Pearson) is obtained as described in Section~\ref{sec:supervised_fine_tuning} by using the $\mathcal{L}_{\text{corr}}$ loss defined in Equation~\eqref{eq:pearson_correlation} during the multi-task learning for STS datasets.
The second variant instead uses a standard mean squared error (MSE) loss, which follows the work of~\citet{reimers2019sentence}.
Additionally, there is a third variant that does not make use of the STS data. 
The STS data consists of rows with two text values complemented by scores in different ranges between 0 and 5.
Since cosine similarity values can not express scores beyond $1.0$, we re-scaled those scores to fit on the interval $[0, 1]$.
%
% This subsection investigates how including STS training data during stage~III via the discussed multitask learning approach impacts performance on downstream STS tasks. To this end, we study three model variants based on the Spanish and German pair models (i.e., the embedding models obtained after stage~II). \todo{describe the pearson model, depends on decision which model we describe in rest of paper}. The second model, which we will refer to as the \emph{MSE}-variant, ablates the scale-invariance of our proposed Pearson loss; instead of using the Pearson sample correlation as a loss function on the STS training sets, we follow~\citet{reimers2019sentence} by utilizing the mean squared error between predicted cosine similarities and ground-truth relevance judgements. The ground-truth relevance labels are scaled linearly to the interval $[0, 1]$. Our third model variant serves as a simple baseline and differs from the previous approaches in that no STS training data is used at all during stage~III.
%
%
% Table~\ref{tab:ablation_study_sts} shows the Spearman correlation between the ground truth similarity scores of the STS12 test set~\cite{agirre2012semeval} and the cosine similarity scores produced by the model variants.
Table~\ref{tab:ablation_study_sts} shows the Spearman correlation between the ground truth similarity scores of various STS test datasets and the cosine similarity scores produced by the model variants.
Both in the German and Spanish settings, we observe that the Pearson variant outperforms the other two models.
The effect is especially strong for SICK-R~\cite{marelli-etal-2014-sick} and STS12~\cite{agirre2012semeval}, where the corresponding STS datasets are included in our training data and less visible for out-of-domain evaluation.
While the Pearson sample correlation loss seems most suitable for incorporating STS training data into stage~\ref{stage:three}, we also find that using a simple MSE regression loss is beneficial over using no STS data at all.
\section{Conclusion}
\label{sec:conclusion}
This paper introduces a set of state-of-the-art bilingual models that support English and one other target language.
Alongside this, we compose benchmarks for German and Spanish embedding models and integrate them into the Massive Text Embedding Benchmark (MTEB) to foster research on German and Spanish text embedding models.
Our evaluation results show that those models achieve superior or competitive results compared to related multilingual models while having a lower number of parameters.
For training these models, we apply a novel multi-task learning objective that specifically improves the models' performance on semantic text similarity tasks.
Moreover, we conduct ablation studies to verify our hypothesis that bilingual models achieve better results than multilingual models when fine-tuned on an embedding task.

% We have introduced \JEmbeddingVTwo, a novel embedding model based on a modified BERT architecture. This model eschews positional embeddings and instead employs bi-directional ALiBi slopes to capture positional information. By training a series of embedding models with this innovative architecture on the Web document corpus C4 and subsequently fine-tuning them, we have enabled the encoding of the semantics of both short and long textual values into meaningful vector representations. This effort has produced a new suite of open-source embedding models capable of encoding texts containing up to $8192$ tokens. These embeddings signify a 16x increase in the maximum sequence length compared to leading open-source embedding models. Additionally, our model suite exhibits competitive performance on the MTEB benchmark. We also demonstrate how utilizing extended sequence lengths can offer our models an advantage over those without such capabilities.

\FloatBarrier

\bibliographystyle{unsrtnat}
\bibliography{references}  %%% Uncomment this line and comment out the ``thebibliography'' section below to use the external .bib file (using bibtex).

\balance

\clearpage
\pagenumbering{gobble}
\onecolumn

\appendix
\section{Appendix}
\begin{table*}[ht]
    \centering
    \small{\begin{tabular}{lcc}
\toprule
 & \JinaEs & \MultilingualEFiveBase \\
\midrule
\bfseries Retrieval Average [nDCG@10] & \bfseries 0.511 & 0.478 \\
MIRACLRetrieval (es) & 0.809 & \bfseries 0.820 \\
MintakaESRetrieval (es) & 0.283 & \bfseries 0.299 \\
SpanishPassageRetrievalS2P (es) & \bfseries 0.421 & 0.394 \\
SpanishPassageRetrievalS2S (es) & \bfseries 0.704 & 0.654 \\
XMarketES (es) & \bfseries 0.197 & 0.119 \\
XPQAESRetrieval (es) & \bfseries 0.654 & 0.583 \\
\midrule
\bfseries Clustering Average [v-measure]& \bfseries 0.440 & 0.413 \\
FloresClusteringS2S (es) & \bfseries 0.398 & 0.367 \\
SpanishNewsClusteringP2P (es) & \bfseries 0.483 & 0.458 \\
\midrule
\bfseries Reranking Average [mAP]& \bfseries 0.739 & 0.736 \\
MIRACL (es) & \bfseries 0.739 & 0.736 \\
\midrule
\bfseries Classification Average [acc.]& 0.671 & \bfseries 0.685 \\
AmazonReviewsClassification (es) & 0.387 & \bfseries 0.405 \\
MTOPDomainClassification (es) & 0.899 & \bfseries 0.906 \\
MTOPIntentClassification (es) & 0.688 & \bfseries 0.713 \\
MassiveIntentClassification (es) & 0.669 & \bfseries 0.684 \\
MassiveScenarioClassification (es) & 0.712 & \bfseries 0.715 \\
\midrule
\bfseries Pair-Classification Average [AveP] & \bfseries 0.590 & 0.551 \\
PawsX (es) & \bfseries 0.590 & 0.551 \\
\midrule
\bfseries STS Average [Spearman] & \bfseries 0.788 & 0.783 \\
STS17 (es) & \bfseries 0.882 & 0.867 \\
STS22 (es) & \bfseries 0.680 & 0.666 \\
STSES (es) & 0.802 & \bfseries 0.816 \\
%\hline
% \bfseries Overall Average &  &  \\ % Let's not do this please
\bottomrule
\end{tabular}}
    \caption{Performance of \JinaEs~and \MultilingualEFiveBase~on Spanish MTEB tasks.}
    \label{table:spanish_mteb}
\end{table*}

\begin{table*}[ht]
    \centering
    \small{
        \begin{tabular}{lcc}
\toprule
 & \JinaDe & \MultilingualEFiveBase \\
\midrule
\bfseries Retrieval Average [nDCG@10]& \bfseries 0.387 & 0.342 \\
GerDaLIR (de) & \bfseries 0.172 & 0.069 \\
GermanDPR (de) & \bfseries 0.795 & \bfseries 0.795 \\
XMarket (de) & \bfseries 0.195 & 0.163 \\
\midrule
\bfseries Clustering Average [v-measure]& 0.299 & \bfseries 0.328 \\
BlurbsClusteringP2P (de) & 0.349 & \bfseries 0.394 \\
BlurbsClusteringS2S (de) & \bfseries 0.167 & 0.157 \\
TenKGnadClusteringP2P (de) & 0.431 & \bfseries 0.439 \\
TenKGnadClusteringS2S (de) & 0.250 & \bfseries 0.320 \\
\midrule
\bfseries Reranking Average [mAP] & 0.639 & \bfseries 0.648 \\
MIRACL (de) & 0.639 & \bfseries 0.648 \\
\midrule
\bfseries Classification Average [acc.] & 0.665 & \bfseries 0.687 \\
AmazonCounterfactualClassification (de) & 0.692 & \bfseries 0.717 \\
AmazonReviewsClassification (de) & 0.385 & \bfseries 0.418 \\
MTOPDomainClassification (de) & 0.892 & \bfseries 0.896 \\
MTOPIntentClassification (de) & 0.653 & \bfseries 0.712 \\
MassiveIntentClassification (de) & 0.650 & \bfseries 0.661 \\
MassiveScenarioClassification (de) & 0.718 & \bfseries 0.720 \\
\midrule
\bfseries Pair-Classification Average [AveP] & \bfseries 0.583 & 0.541 \\
PawsX (de) & \bfseries 0.583 & 0.541 \\
\midrule
\bfseries STS Average [Spearman] & \bfseries 0.714 & 0.674 \\
GermanSTSBenchmark* (de) & \bfseries 0.845 & 0.789 \\
STS22 (de) & \bfseries 0.583 & 0.560 \\
\bottomrule
\end{tabular}
        \begin{minipage}{0.71\textwidth}\raggedright
        \vspace{2mm}
        * Contains some STS train data
        \end{minipage}
    }
    \caption{Performance of \JinaDe~and \MultilingualEFiveBase~on German MTEB tasks.}
    \label{table:german_mteb}
\end{table*}

\begin{table*}
    \centering
    \small{\begin{tabular}{lcc}
\toprule
     & \JinaEs & \MultilingualEFiveBase  \\
     \midrule
\bfseries STS Average [Spearman] & \bfseries 0.826 & 0.752 \\
STS17 (es-en) & \bfseries 0.865 & 0.765 \\
STS22 (es-en) & \bfseries 0.788 & 0.740 \\
\bottomrule

\end{tabular}}
    \caption{Performance of \JinaEs~and \MultilingualEFiveBase~on Spanish-English cross-lingual tasks}
    \label{table:spanish_english_mteb}
\end{table*}

\begin{table*}
    \centering
    \small{\begin{tabular}{lcc}
\toprule
 & \JinaDe & \MultilingualEFiveBase \\
\midrule
\bfseries STS Average [Pearson] & \bfseries 0.718  & 0.685 \\
STS17 (en-de) & \bfseries 0.865 & 0.821 \\
STS22 (de-en) & \bfseries 0.570 & 0.549 \\
\bfseries Bitext Mining Average [F1] & 0.990 & \bfseries 0.991 \\
BUCC (de-en) & 0.990 & \bfseries 0.991 \\
\bottomrule
\end{tabular}
}
    \caption{Performance of \JinaDe~and \MultilingualEFiveBase~on German-English cross-lingual tasks}
    \label{table:german_english_mteb}
\end{table*}
\begin{table*}[ht]
    \centering
    \vspace{-1.3cm}
    \small{
        \begin{tabular}{lccc}
\toprule
 & \JinaDe & \JinaEs & \MultilingualEFiveBase \\
\midrule
\bfseries Retrieval Average [mDCG@10] & 0.441 & 0.464 & \bfseries 0.489 \\
ArguAna (en) & 0.495 & \bfseries 0.501 & 0.442 \\
CQADupstackRetrieval (en) & 0.368 & \bfseries 0.388 & 0.385 \\
ClimateFEVER (en) & 0.222 & \bfseries 0.271 & 0.239 \\
DBPedia (en) & 0.320 & 0.326 & \bfseries 0.404 \\
FEVER (en) & 0.712 & 0.784 & \bfseries 0.794 \\
FiQA2018 (en) & 0.339 & 0.370 & \bfseries 0.382 \\
HotpotQA (en) & 0.548 & 0.595 & \bfseries 0.686 \\
MSMARCO (en) & 0.345 & 0.355 & \bfseries 0.423 \\
NFCorpus (en) & 0.280 & 0.303 & \bfseries 0.325 \\
NQ (en) & 0.481 & 0.507 & \bfseries 0.600 \\
QuoraRetrieval (en) & 0.877 & \bfseries 0.881 & 0.876 \\
SCIDOCS (en) & 0.167 & \bfseries 0.174 & 0.172 \\
SciFact (en) & 0.609 & 0.627 & \bfseries 0.693 \\
TRECCOVID (en) & 0.648 & 0.674 & \bfseries 0.698 \\
Touche2020 (en) & 0.205 & 0.198 & \bfseries 0.214 \\
\midrule
\bfseries Clustering Average [v-measure] & 0.403 & \bfseries 0.405 & 0.400 \\
ArxivClusteringP2P (en) & \bfseries 0.419 & 0.415 & 0.403 \\
ArxivClusteringS2S (en) & 0.325 & 0.322 & \bfseries 0.354 \\
BigPatentClustering (en) & 0.230 & 0.204 & \bfseries 0.240 \\
BiorxivClusteringP2P (en) & 0.344 & \bfseries 0.352 & 0.350 \\
BiorxivClusteringS2S (en) & 0.290 & 0.289 & \bfseries 0.295 \\
MedrxivClusteringP2P (en) & 0.316 & \bfseries 0.322 & 0.289 \\
MedrxivClusteringS2S (en) & \bfseries 0.294 & 0.288 & 0.284 \\
RedditClustering (en) & 0.479 & \bfseries 0.483 & 0.424 \\
RedditClusteringP2P (en) & 0.549 & \bfseries 0.560 & 0.552 \\
StackExchangeClustering (en) & \bfseries 0.555 & 0.554 & 0.553 \\
StackExchangeClusteringP2P (en) & 0.328 & \bfseries 0.336 & 0.305 \\
TwentyNewsgroupsClustering (en) & \bfseries 0.405 & 0.402 & 0.360 \\
WikiCitiesClustering (en) & 0.703 & 0.733 & \bfseries 0.796 \\
reranking average & 0.549 & \bfseries 0.553 & 0.548 \\
AskUbuntuDupQuestions (en) & \bfseries 0.612 & \bfseries 0.612 & 0.582 \\
MindSmallReranking (en) & 0.303 & 0.306 & \bfseries 0.310 \\
SciDocsRR (en) & 0.786 & 0.793 & \bfseries 0.807 \\
StackOverflowDupQuestions (en) & 0.496 & \bfseries 0.502 & 0.494 \\
\midrule
\bfseries Classification Average [acc.] & 0.688 & 0.690 & \bfseries 0.730 \\
AmazonCounterfactualClassification (en) & 0.740 & 0.743 & \bfseries 0.790 \\
AmazonPolarityClassification (en) & 0.802 & 0.783 & \bfseries 0.906 \\
AmazonReviewsClassification (en) & 0.395 & 0.383 & \bfseries 0.445 \\
Banking77Classification (en) & 0.839 & \bfseries 0.853 & 0.827 \\
EmotionClassification (en) & 0.452 & \bfseries 0.466 & 0.452 \\
ImdbClassification (en) & 0.699 & 0.675 & \bfseries 0.855 \\
MTOPDomainClassification (en) & 0.903 & 0.904 & \bfseries 0.931 \\
MTOPIntentClassification (en) & 0.686 & 0.720 & \bfseries 0.753 \\
MassiveIntentClassification (en) & 0.696 & 0.708 & \bfseries 0.721 \\
MassiveScenarioClassification (en) & 0.741 & 0.738 & \bfseries 0.771 \\
ToxicConversationsClassification (en) & 0.695 & \bfseries 0.700 & 0.698 \\
TweetSentimentExtractionClassification (en) & \bfseries 0.613 & 0.611 & \bfseries 0.613 \\
\midrule
\bfseries Pair-Classification [AveP] & 0.835 & \bfseries 0.846 & 0.836 \\
SprintDuplicateQuestions (en) & 0.940 & \bfseries 0.956 & 0.930 \\
TwitterSemEval2015 (en) & 0.705 & 0.716 & \bfseries 0.722 \\
TwitterURLCorpus (en) & 0.860 & \bfseries 0.864 & 0.855 \\
\midrule
\bfseries STS Average [Spearman] & 0.820 & \bfseries 0.835 & 0.803 \\
BIOSSES (en) & 0.780 & 0.830 & \bfseries 0.851 \\
SICK-R (en) & 0.811 & \bfseries 0.830 & 0.785 \\
STS12 (en) & 0.801 & \bfseries 0.816 & 0.767 \\
STS13 (en) & 0.851 & \bfseries 0.868 & 0.780 \\
STS14 (en) & 0.811 & \bfseries 0.836 & 0.766 \\
STS15 (en) & 0.878 & \bfseries 0.887 & 0.882 \\
STS16 (en) & 0.836 & \bfseries 0.860 & 0.843 \\
STS17 (en) & \bfseries 0.893 & 0.886 & 0.878 \\
STS22 (en) & \bfseries 0.682 & 0.656 & 0.618 \\
STSBenchmark* (en) & 0.862 & \bfseries 0.878 & 0.856 \\
\midrule
\bfseries Summarization Average [Spearman] & \bfseries 0.318 & 0.299 & 0.301 \\
SummEval (en) & \bfseries 0.318 & 0.299 & 0.301 \\
\bottomrule
\end{tabular}

        \begin{minipage}{0.89\textwidth}\raggedright
        \vspace{2mm}
        * Contains some STS train data
        \end{minipage}
    }
    \caption{Performance of Jina models and \MultilingualEFiveBase~on English MTEB tasks.}
    \label{table:english_mteb}
\end{table*}





% \end{thebibliography}

\end{document}